\pdfoutput=1

\documentclass[11pt]{article}

\usepackage{acl}

\usepackage{times}
\usepackage{latexsym}
\usepackage{graphicx}
\usepackage[inkscapearea=page]{svg}
\usepackage{tabularx}
\usepackage{hyperref}
\usepackage{multirow}
\usepackage{bm}
\usepackage{soul}

\usepackage[T1]{fontenc}

\usepackage[utf8]{inputenc}

\usepackage{microtype}

\usepackage{inconsolata}

\title{\textit{QASE} Enhanced PLMs: Improved Control in Text Generation for MRC}

\author{Lin Ai\quad Zheng Hui\quad Zizhou Liu\quad Julia Hirschberg  \\
       Columbia University, New York, NY \\ 
         \texttt{\{lin.ai, julia\}@cs.columbia.edu} \\
         \texttt{\{zh2483, zl2889@columbia.edu\}@columbia.edu} }

\begin{document}
\maketitle
\begin{abstract}
To address the challenges of out-of-control generation in generative models for machine reading comprehension (MRC), we introduce the \textbf{Q}uestion-\textbf{A}ttended \textbf{S}pan \textbf{E}xtraction (\textit{QASE}) module. Integrated during the fine-tuning of pre-trained generative language models (PLMs), \textit{QASE} enables these PLMs to match SOTA extractive methods and outperform leading LLMs like GPT-4 in MRC tasks, without significant increases in computational costs. \footnote{Our code is available at \href{https://anonymous.4open.science/r/QASE-7753/README.md}{this anonymous repo link}.}
\end{abstract}

\section{Introduction}
Machine Reading Comprehension (MRC) is a critical NLP challenge. Mainstream approaches to MRC extract a relevant piece of text from the context in response to a question \cite{wang-etal-2018-multi-granularity, yan2019deep, chen2020question}, but in real-world application, the correct answers often span multiple passages or are implicit \cite{li-etal-2021-addressing-semantic}. Exploring generative models, in addition to extractive methods, is essential.

Generative models, however, underperform in MRC due to out-of-control generation \cite{li-etal-2021-addressing-semantic}. This leads to two main challenges: (1) ill-formed generated answers, containing incomplete or redundant phrases, and (2) factual inconsistency in the generated answers deviating from the correct response. In this paper, we address these by introducing a lightweight \textbf{Q}uestion-\textbf{A}ttended \textbf{S}pan \textbf{E}xtraction (\textit{QASE}) module. We fine-tune multiple open-source generative pre-trained language models (PLMs) on various MRC datasets to assess the module's efficacy in guiding answer generation. Our contributions include: \textbf{(1)} Developing \textit{QASE} to improve fine-tuned generative PLMs' quality and factual consistency on MRC tasks, matching SOTA extractive methods and surpassing GPT-4; \textbf{(2)} \textit{QASE} boosts performance without significantly increasing computational costs, benefiting researchers with limited resources.

\section{Related Work}
\label{sec:related_work}

Most \textbf{current studies on MRC} involve predicting the start and end positions of the answer spans from a given context \cite{ohsugi-etal-2019-simple, lan2019albert, bachina-etal-2021-ensemble, chen2022good} using encoder-only PLM models such as BERT and XLM-Roberta. To handle the multi-span setting, some studies frame the problem as a sequence tagging task \cite{segal-etal-2020-simple}, and others explore ways to combine models with different tasks \cite{hu-etal-2019-multi, lee2023liquid, zhang2023many}. While these extractive-based methods mainly utilize encoder-only models, there is also research focuses on using generative language models \cite{yang2020multi, li-etal-2021-addressing-semantic, su-etal-2022-read}. 

\textbf{Retrieval-augmented text generation} (RAG) augments the input of PLMs with in-domain \cite{gu2018search, weston-etal-2018-retrieve, saha-srihari-2023-argu} or external knowledge \cite{su2021prototype, xiao-etal-2021-transductive} to control the quality and factual consistency of generated content. It has become a new text generation paradigm in many NLP tasks \cite{li2022survey}, such as dialogue response generation \cite{wu2021controllable, liu-etal-2023-recap} and machine translation \cite{he-etal-2021-fast, zhu-etal-2023-ink}. However, not much work focuses on selective MRC. Our approach diverges from RAG as it directly fine-tunes the weights of the PLMs rather than altering the input to the PLMs with additional information.

\section{Method}


\textbf{\underline{\textit{Question-Attended Span Extraction}}} To guide text generation, we use  \textit{QASE}, a question-attended span extraction module, during fine-tuning the generative PLMs. \textit{QASE} focuses  model attention on potential answer spans within the original context. We cast span extraction as a sequence tagging problem and employ the Inside-Outside (IO) tagging schema, where each sequence  token is tagged as `inside' (\textbf{\textit{I}}) if part of a relevant span, or `outside' (\textbf{\textit{O}}) if not. This schema works well for both single- and multi-span extraction settings, achieving comparable or even better performance than the well-known BIO tagging format \cite{huang2015bidirectional}, as shown by \citet{segal-etal-2020-simple}.

\vspace{-0.2cm}
\begin{figure}[!htbp]
    \centering
    \includegraphics[width=0.44\textwidth]{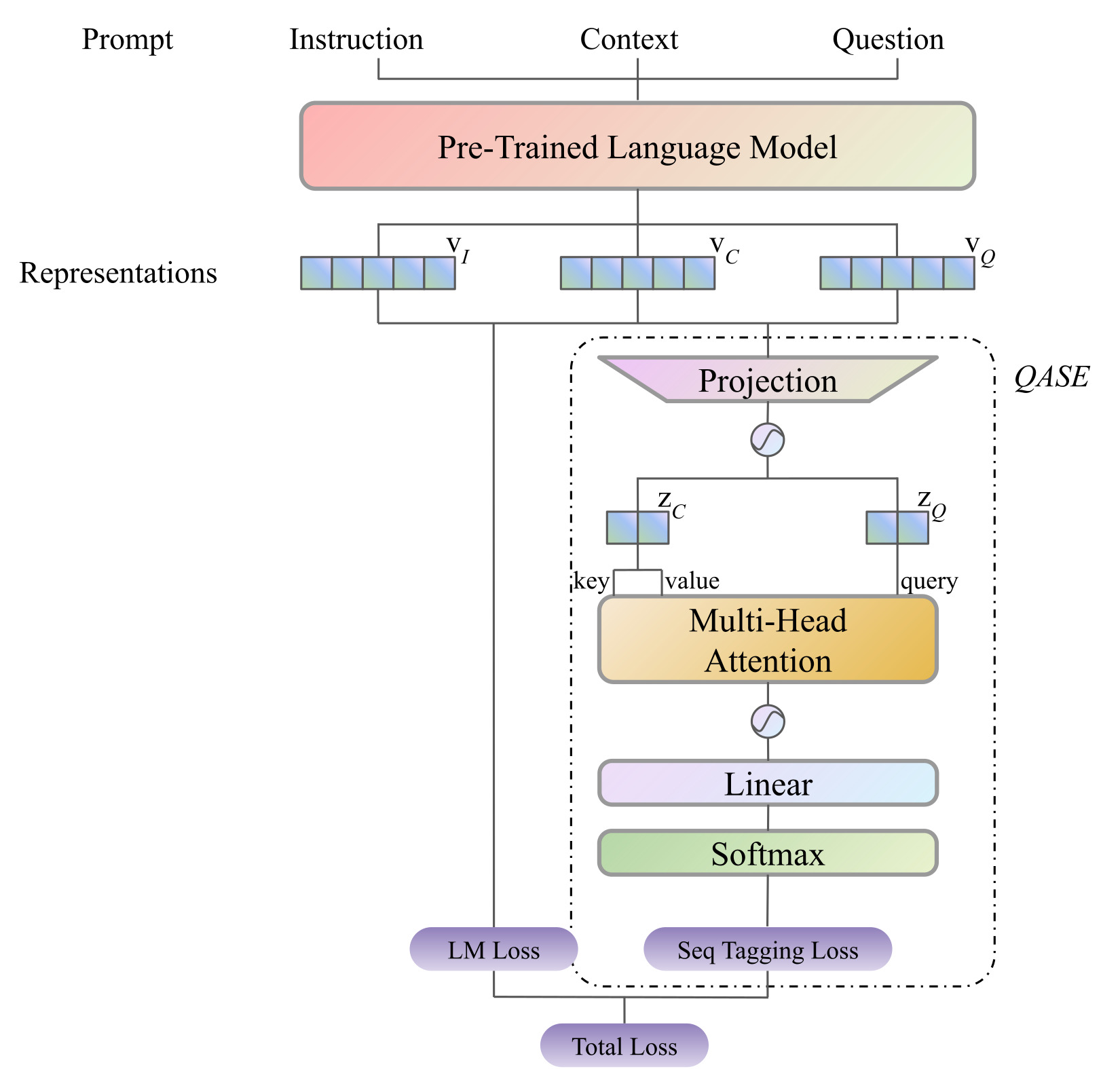}
    \caption{\textit{QASE}-enhanced model architecture}
    \label{fig:model}
    \vspace{-0.2cm}
\end{figure}

The architecture of our model is shown in Figure \ref{fig:model}. An input context and question pair and an instruction are first tokenized and fed into the PLM. The hidden states output from the PLM is then passed through projection layers to produce embeddings $z_i = ReLU(W_{proj}v_i + b_{proj})$, where $v_i \in R^d$ is the PLM output hidden state of the $i^{th}$ token.

To learn context tokens representations in relation to specific questions, we employ a \textbf{multi-head attention} mechanism (\textit{MHA}). Each head in \textit{MHA} focuses to different aspects of the context as it relates to the question, using question embeddings as the query and context embeddings as key-value pairs. This mechanism aligns the context token representations with the specifics of the queried question. The projected embeddings $z_i$ are passed through \textit{MHA}, and subsequently channeled through a linear layer and a softmax layer to compute $p_i = softmax(W_{lin} \cdot MHA(z_i) + b_{lin})$, which denotes the probability of the $i^{th}$ token being inside the answer spans. We then compute the sequence tagging loss using the cross entropy loss $L_{QASE} = -\frac{1}{N} \sum_{i=1}^N \sum_{j=0}^1 y_{ij}log(p_{ij})$, where $j \in {0, 1}$ corresponds to class \textbf{\textit{O}} and class \textbf{\textit{I}}, and $y_{ij}$ is a binary value indicating whether the $i^{th}$ token belongs to class $j$.

\textbf{\underline{\textit{Fine-Tuning and Inference}}} We fine-tune the PLMs using multi-task learning, simultaneously optimizing both the language modeling loss and sequence tagging loss: $L = L_{LML} + \beta L_{QASE}$, where $\beta$ is a hyper-parameter that controls the weight of the span extraction task. This approach enhances the PLMs' ability to generate answers well-founded in the context and relevant answer spans. During inference, only the generation component of the fine-tuned model is employed.

\section{Experiments}
\label{sec:experiments}

\begin{table*}[htbp!]
    \centering
    \small
    \begin{tabular}{cr||c|c|c|c|c}
    \hline\hline
                                     & \textbf{} & \textbf{Llama2} & \textbf{Alpaca} & \textbf{Flan-T5-Small} & \textbf{Flan-T5-Base} & \textbf{Flan-T5-Large} \\\hline\hline
    \textbf{SQuAD}     & no \textit{QASE}   & 36.68 | 47.06          & 27.88 | 43.95          & 77.33 | 85.51          & 82.09 | 89.56          & 83.16 | 90.71          \\
    (EM | F1)                       & \textit{QASE} & \textbf{37.22} | \textbf{47.69} & \textbf{37.31} | \textbf{47.62} & \textbf{77.66} | \textbf{85.90} & \textbf{82.20} | \textbf{90.24} & \textbf{84.13} | \textbf{91.70} \\\hline
    \textbf{MultiSpanQA} & no \textit{QASE}   & 50.93 | 68.14          & \textbf{52.73} | 69.10          & \textbf{59.13} | 76.49          & 64.66 | 81.41          & \textbf{67.41} | 83.09          \\
    (EM F1 | Overlap F1)          & \textit{QASE}      & \textbf{51.75} | \textbf{70.39}   & 52.20 | \textbf{70.01}   & 59.08 | \textbf{77.10}      & \textbf{64.87} | \textbf{81.50}      & 66.92 | \textbf{84.22}      \\\hline
    \textbf{Quoref} & no \textit{QASE}        & 45.52 | 52.09   & -               & 58.21 | 63.30      & 72.77 | 80.90      & 75.17 | 80.49      \\
    (EM | F1)                       & \textit{QASE}      & \textbf{54.28} | \textbf{60.44}   & -               & \textbf{60.70} | \textbf{66.88}      & \textbf{75.17} | \textbf{81.18}      & \textbf{76.19} | \textbf{82.13}   \\\hline\hline  
    \end{tabular}
    \caption{Performance of fine-tuned PLMs with or without \textit{QASE} on each dataset.}
    \label{tab:results}
    \vspace{-0.2cm}
\end{table*}

\textbf{\underline{\textit{Datasets and Metrics}}} We utilize these 3 MRC datasets.
\textbf{(1) SQuAD} \cite{rajpurkar-etal-2016-squad}: A benchmark dataset consisting of 100K+ questions with single-span answers. We use SQuAD v1.1. Since the official evaluation on v1.1 has long been ended, we report our results on the official v1.1 development set.
\textbf{(2) MultiSpanQA} \cite{li-etal-2022-multispanqa}: This dataset consists of over 6.5k question-answer pairs. Unlike most existing single-span answer MRC datasets, MultiSpanQA focuses on multi-span answers.
\textbf{(3) Quoref} \cite{dasigi-etal-2019-quoref}: A benchmark dataset containing more than 24K questions, with most answers being single-span and  $\sim$10\% being multi-span.
Following the conventions of the datasets' official leaderboards (listed in \ref{subsec:leaderboard}), we employ exact match (EM) and partial match (Overlap) F1 scores as metrics on MultiSpanQA, and exact match percentage and macro-averaged F1 score on SQuAD and Quoref.

\textbf{\underline{\textit{Experimental Setup}}} To evaluate the effectiveness of \textit{QASE} independent of any specific language model, we experiment with multiple open-source LLMs. These include both decoder-only LLMs, such as Llama 2 \cite{touvron2023llama} and Alpaca \cite{alpaca}, and an encoder-decoder model, Flan-T5 \cite{chung2022scaling}. For Llama 2 and Alpaca, we fine-tune the pre-trained 7B version using LoRA \cite{hu2021lora} and instruction-tuning (see \ref{subsec:prompts} for instruction templates). For Flan-T5 family models, we fine-tune the small, the base, and the large versions. The trainable parameters for each model is provided in Table \ref{tab:params}.

\begin{table}[!htbp]
    \centering
    \small
    
    \begin{tabular}{c||ccc}
    \hline\hline
                           & \multicolumn{3}{c}{\textbf{Trainable Parameters}} \\
    \textbf{}              & no \textit{QASE}       & \textit{QASE}         & $\Delta$params       \\\hline\hline
    \textbf{\begin{tabular}[c]{@{}c@{}}Llama2/Alpaca\\with LoRA\end{tabular}} & 4.2M & 7.3M & 3.1M \\\hline
    \textbf{Flan-T5-Small} & 77.0M           & 78.2M          & 1.3M           \\\hline
    \textbf{Flan-T5-Base}  & 247.6M          & 248.9M         & 1.4M           \\\hline
    \textbf{Flan-T5-Large} & 783.2M          & 784.7M         & 1.5M           \\\hline\hline
    \end{tabular}
    \caption{Trainable parameters of experimented models.}
    \label{tab:params}
    \vspace{-0.5cm}
\end{table}

We set the hyper-parameters $\beta = 1$ and the learning rate $lr = 1e-4$. For LoRA fine-tuning applied to Llama 2 and Alpaca models, we specify a rank $r = 8$, $\alpha = 32$, and a dropout rate of $0.05$. The methodology for selecting these hyper-parameters is detailed in \ref{subsec:hp_selection}. We train all our models on single GPUs, using a batch size of 2-4 depending on the VRAM of the respective GPUs. We use four types of GPUs: A40, A10, A5500, and A100. Models are trained for 3 epochs or until convergence. 



\textbf{\underline{\textit{Experiment Results}}} To evaluate the efficacy of the \textit{QASE}, we examine the performance of various PLMs fine-tuned with and without \textit{QASE}, as shown in Table \ref{tab:results}. Generally, models fine-tuned with \textit{QASE} outperform those fine-tuned without it. In particular, for SQuAD, \textit{QASE}-enhanced model demonstrate an EM percentage increase of up to 33.8\% and an F1 score upsurge of up to 8.4\% over vanilla fine-tuned models. For MultiSpanQA, there is an improvement of up to 1.6\% in the EM F1 and up to 3.3\% in the overlap F1. Likewise, on Quoref, there is an improvement of up to 19.2\% in the EM percentage and up to 16.0\% in the F1 score. These results show that, by employing \textit{QASE}, generative-based PLMs can be fine-tuned to produce well-formed, context-grounded, and better-quality answers in MRC tasks compared to the vanilla fine-tuning approach. For reference, we also compare the fine-tuned PLMs to their corresponding PLMs in zero-shot settings, as presented in Appendix \ref{subsec:full_results}.

\textbf{\underline{\textit{Computational Costs}}} Table \ref{tab:params} shows that integrating \textit{QASE} slightly raises the number of trainable parameters in PLMs, with the increase dependent on the models' hidden sizes. Significantly, for the largest model, Flan-T5-Large, \textit{QASE} adds just 0.2\% more parameters, indicating that \textit{QASE} enhances the capabilities of fine-tuned PLMs in MRC without major increase in computational resources.

\textbf{\underline{\textit{Model Comparisons}}} Our top model, Flan-T5-Large$_{QASE}$, is further benchmarked against leading models on each dataset's official leaderboard, alongside zero-shot GPT-3.5-Turbo and GPT-4. GPT-3.5-Turbo stands as one of OpenAI's most efficient models in terms of capability and cost, while GPT-4 shows superior reasoning abilities \cite{liu-etal-2023-system}. Studies indicate their superiority over traditional fine-tuning methods in most logical reasoning benchmarks \cite{liu2023evaluating}. The prompts used to query the GPT variants are detailed in Appendix \ref{subsec:prompts}.
On SQuAD, as showed in Table \ref{tab:squad_gpt}, Flan-T5-Large$_{QASE}$ surpasses human performance, equaling the NLNet model. 
Additionally, it surpasses GPT-4 by 113.8\% on the exact match score and 32.6\% on F1.
\begin{table}[!htbp]
    \centering
    \small
    \begin{tabular}{l||cc}
    \hline\hline
    \textbf{}          & \textbf{EM}     & \textbf{F1} $\uparrow$     \\ \hline\hline
    GPT-3.5-Turbo      & 36.944          & 65.637          \\
    GPT-4              & 39.347          & 69.158          \\
    Human Performance  & 82.304          & 91.221          \\
    BERT-Large \cite{devlin-etal-2019-bert} & 84.328 & 91.281 \\
    MSRA NLNet (ensemble)   & \textbf{85.954} & 91.677      \\ \hline
    Flan-T5-Large$_{QASE}$ & 84.125 & \textbf{91.701} \\ \hline\hline
    \end{tabular}
    \caption{Flan-T5-Large$_{QASE}$ and baselines on \textbf{SQuAD}.}
    \label{tab:squad_gpt}
    \vspace{-0.3cm}
\end{table}
On MultiSpanQA, Table \ref{tab:multispanqa_gpt} shows that Flan-T5-Large$_{QASE}$ outperforms LIQUID \cite{lee2023liquid}, which currently ranks \#1 on the leaderboard, with respect to the overlap F1 score. Moreover, it surpasses GPT-4 by 4.5\% on the exact match F1 and 1.5\% on the overlap F1.
\begin{table}[!htbp]
    \centering
    \small
    \begin{tabular}{l||cc}
    \hline\hline
    \textbf{}                   & \textbf{EM F1}  & \textbf{Overlap F1} $\uparrow$ \\ \hline\hline
    GPT-3.5-Turbo               & 59.766          & 81.866              \\
    GPT-4                       & 64.027          & 82.731              \\ 
    LIQUID \cite{lee2023liquid} & \textbf{73.130} & 83.360              \\ \hline
    Flan-T5-Large$_{QASE}$      & 66.918          & \textbf{84.221}     \\ \hline\hline
    \end{tabular}
    \caption{Performance of Flan-T5-Large$_{QASE}$ and baselines on \textbf{MultiSpanQA}.}
    \label{tab:multispanqa_gpt}
    \vspace{-0.3cm}
\end{table}
\begin{table}[!htbp]
    \centering
    \small
    \begin{tabular}{l||cc}
    \hline\hline
    \textbf{}          & \textbf{EM}    & \textbf{F1} $\uparrow$    \\ \hline\hline
    GPT-3.5-Turbo      & 50.22          & 59.51          \\
    GPT-4              & 68.07          & 78.34          \\
    CorefRoberta-Large \cite{ye-etal-2020-coreferential} & 75.80  & \textbf{82.81} \\ \hline
    Flan-T5-Large$_{QASE}$ & \textbf{76.19} & 82.13          \\ \hline\hline
    \end{tabular}
    \caption{Performance of Flan-T5-Large$_{QASE}$ and baselines on \textbf{Quoref}.}
    \label{tab:quoref_gpt}
    \vspace{-0.3cm}
\end{table}
On Quoref, Table \ref{tab:quoref_gpt} shows that Flan-T5-Large$_{QASE}$ is comparable to CorefRoberta-Large \cite{ye-etal-2020-coreferential}, which ranks \#9 on the leaderboard, with a 0.5\% higher exact match. Furthermore, it outperforms GPT-4 by 11.9\% on the exact match and 4.8\% on F1.

All top-performing models on these datasets' leaderboards, equaling or exceeding Flan-T5-Large$_{QASE}$, are encoder-only extractive models. Therefore, these results demonstrate that \textit{QASE}-enhanced generative PLMs can be fine-tuned to match or exceed the capabilities of SOTA extractive models and outperform leading LLMs in MRC.

\textbf{\underline{\textit{Ablation Studies}}} To demonstrate the superiority of the \textit{QASE} architecture, we compared Flan-T5-Large$_{QASE}$ with vanilla fine-tuned Flan-T5-Large$_{FT}$ and Flan-T5-Large$_{baseline}$. The baseline span extraction module lacks the \textit{MHA} component, making it a standard architecture for fine-tuning pre-trained encoders for downstream sequence tagging tasks. We also explored both question-first (\textit{qf}) and context-first prompting strategies, with further details and analysis provided in Appendix \ref{subsec:ablation_details}, where the model architecture is also illustrated. 

Table \ref{tab:ablation} shows that the baseline-embedded model performs better with a question-first prompting strategy, as Flan-T5-Large$_{baseline_{qf}}$ surpasses Flan-T5-Large$_{baseline}$ and Flan-T5-Large$_{FT_{qf}}$. Conversely, the baseline span extraction module decreases performance in context-first prompting, where Flan-T5-Large$_{baseline}$ underperforms compared to Flan-T5-Large$_{FT}$. This suggests that adding an auxiliary span extraction module without careful design can negatively affect instruction fine-tuning. Meanwhile, the \textit{QASE}-enhanced model excels over both vanilla fine-tuned and baseline-embedded models in both prompting scenarios, demonstrating its architectural superiority. Specifically, in context-first setting, Flan-T5-Large$_{QASE}$ significantly outperforms Flan-T5-Large$_{baseline}$ with a 4.3\% higher F1. 


\begin{table}[!htbp]
    \centering
    \small
    \begin{tabular}{c||cc}
    \hline\hline
             & \textbf{EM}     & \textbf{F1} $\uparrow$    \\\hline\hline
    Flan-T5-Large$_{baseline}$ & 79.877 & 87.918\\
    Flan-T5-Large$_{FT_{qf}}$  & 80.378 & 88.176\\
    Flan-T5-Large$_{baseline_{qf}}$ & 81.125          & 89.043          \\
    Flan-T5-Large$_{QASE_{qf}}$     & 81.485          & 89.077          \\
    Flan-T5-Large$_{FT}$ & 83.159 & 90.712 \\\hline
    Flan-T5-Large$_{QASE}$         & \textbf{84.125} & \textbf{91.701}  \\\hline\hline
    \end{tabular}
    \caption{Performance of vanilla, baseline-, and \textit{QASE}-enhanced fine-tuned Flan-T5-Large on \textbf{SQuAD}.}
    \label{tab:ablation}
    \vspace{-0.3cm}
\end{table}


\textbf{\underline{\textit{Factual Consistency}}} While token-based EM and F1 scores measure the structural quality of generated text, they do not reflect factual accuracy relative to the context. For this we used $Q^2$ \cite{honovich2021q}, an automatic metric for assessing factual consistency in generated text, which uses question generation and answering methods over token-based matching. We compared fine-tuned Flan-T5-Large with and without \textit{QASE} in both single-span (SQuAD) and multi-span (MultiSpanQA) answer settings. Table \ref{tab:q2} shows that \textit{QASE}-enhanced models consistently outperform the vanilla fine-tuned model. On SQuAD, $Q^2$ NLI score is improved by 1.0\%, and on MultiSpanQA, it is improved by 16.0\%. Beyond the $Q^2$ statistical analysis, our detailed case studies in Appendix \ref{subsec:case_studies} highlight Flan-T5-Large$_{QASE}$'s improved performance. These examples show the model's better alignment with relevant context, its enhanced understanding of complex sentences, its skill in synthesizing answers from dispersed information, and its superior use of pre-existing real-world knowledge in generating answers.

\begin{table}[!htbp]
    \centering
    \small
    \begin{tabular}{cr||cc}
    \hline\hline
                                          & Flan-T5-Large& $\boldsymbol{Q^2}$ \textbf{F1}  & $\boldsymbol{Q^2}$ \textbf{NLI} \\\hline\hline
    \multirow{2}{*}{\textbf{SQuAD}}       & no \textit{QASE}& 42.927          & 44.983          \\
                                          & \textit{QASE}& \textbf{43.624} & \textbf{45.419} \\\hline
    \multirow{2}{*}{\textbf{MultiSpanQA}} & no \textit{QASE}& 32.889          & 31.433          \\
                                          & \textit{QASE}& \textbf{34.732} & \textbf{36.452} \\\hline\hline
    \end{tabular}
    \caption{$Q^2$ scores of fine-tuned Flan-T5-Large with or without $QASE$ on each dataset.}
    \label{tab:q2}
    \vspace{-0.4cm}
\end{table}

\section{Conclusion and Future Work}

In this study, we address out-of-control text generation of generative PLMs in MRC using \textit{QASE}, a lightweight question-attended span extraction module, during the fine-tuning of PLMs. Our experiments show that \textit{QASE}-enhanced PLMs generate better-quality responses with improved formality and factual consistency, matching SOTA extractive models and outperforming GPT-4 by a significant margin on all three MRC datasets. Importantly, \textit{QASE} improves performance without a significant increase in computational costs, benefiting researchers with limited resources.

In the future, we plan to test our model on generative MRC datasets \cite{nguyen2016ms} to further assess its efficacy in more complex scenarios. Another key focus will be evaluating the model's general ability in answer generation, particularly from the perspective of human perception. This will involve incorporating human annotators in addition to automatic metrics. For a long-term goal, we are looking to expand our work to explore solutions for addressing input- and context-conflicting hallucinations in LLMs.

\clearpage
\section*{Limitations}
\label{sec:limitations}
Due to our limited computational resources, we have been able to perform our experiments on models no larger than Flan-T5-Large. This same constraint led us to only fine-tuning of Llama 2 and Alpaca with LoRA. We note that models based on Llama 2 and Alpaca generally underperform those based on Flan-T5. Apart from the inherent distinctions between decoder-only and encoder-decoder models, and their suitability for different tasks (as seen from the models' zero-shot performance), a possible factor could be the number of trainable parameters during fine-tuning. Specifically, fine-tuning Llama 2 and Alpaca with LoRA results in only 4.2M trainable parameters, while even the smallest Flan-T5 model provides 77.0M trainable parameters, as shown in Table \ref{tab:params}. We acknowledge that many researchers face similar computational resource limitations. Therefore, our research should be very useful, proposing this lightweight module capable of enhancing smaller PLMs to outperform leading LLMs on MRC tasks like these, achieving a balance of effectiveness and affordability.

One foreseeable limitation of our work is the dependency of the fine-tuning process on answer span annotations, since \textit{QASE} works as an auxiliary supervised span extraction module. This reliance on annotated data could potentially limit the model's broader applicability. A prospective exciting future direction to address this limitation is to develop a semi- or unsupervised module that focuses on selecting relevant spans or rationales within a given context. By integrating this module with our current model, we could significantly improve its generalization capabilities, thereby making it more adaptable and effective across a wider range of scenarios.

One popular method to enhance the formality of answers generated by LLMs is through prompt engineering, paired with few-shot or in-context learning techniques. While these strategies offer great advantages, our ultimate goal is to create a system with broad domain generalization, one that minimizes the need for extensive, calibrated prompt engineering and sample selections for task adaptation. Although developing a robust prompt engineering framework or paradigm is an appealing direction, our current focus diverges from this path. As a long-term goal, we aim for a solution that handles diverse tasks with minimal task-specific tuning.

\bibliography{anthology,custom}

\begin{thebibliography}{38}
\expandafter\ifx\csname natexlab\endcsname\relax\def\natexlab#1{#1}\fi

\bibitem[{Bachina et~al.(2021)Bachina, Balumuri, and Kamath~S}]{bachina-etal-2021-ensemble}
Sony Bachina, Spandana Balumuri, and Sowmya Kamath~S. 2021.
\newblock \href {https://doi.org/10.18653/v1/2021.dialdoc-1.9} {Ensemble {ALBERT} and {R}o{BERT}a for span prediction in question answering}.
\newblock In \emph{Proceedings of the 1st Workshop on Document-grounded Dialogue and Conversational Question Answering (DialDoc 2021)}, pages 63--68, Online. Association for Computational Linguistics.

\bibitem[{Chen et~al.(2020)Chen, Xu, Cheng, Xiaochuan, Zhang, Song, Wang, Qi, and Chu}]{chen2020question}
Kunlong Chen, Weidi Xu, Xingyi Cheng, Zou Xiaochuan, Yuyu Zhang, Le~Song, Taifeng Wang, Yuan Qi, and Wei Chu. 2020.
\newblock Question directed graph attention network for numerical reasoning over text.
\newblock \emph{arXiv preprint arXiv:2009.07448}.

\bibitem[{Chen et~al.(2022)Chen, Shou, Gong, and Pei}]{chen2022good}
Nuo Chen, Linjun Shou, Ming Gong, and Jian Pei. 2022.
\newblock From good to best: Two-stage training for cross-lingual machine reading comprehension.
\newblock In \emph{Proceedings of the AAAI Conference on Artificial Intelligence}, volume~36, pages 10501--10508.

\bibitem[{Chung et~al.(2022)Chung, Hou, Longpre, Zoph, Tay, Fedus, Li, Wang, Dehghani, Brahma et~al.}]{chung2022scaling}
Hyung~Won Chung, Le~Hou, Shayne Longpre, Barret Zoph, Yi~Tay, William Fedus, Eric Li, Xuezhi Wang, Mostafa Dehghani, Siddhartha Brahma, et~al. 2022.
\newblock Scaling instruction-finetuned language models.
\newblock \emph{arXiv preprint arXiv:2210.11416}.

\bibitem[{Dasigi et~al.(2019)Dasigi, Liu, Marasovi{\'c}, Smith, and Gardner}]{dasigi-etal-2019-quoref}
Pradeep Dasigi, Nelson~F. Liu, Ana Marasovi{\'c}, Noah~A. Smith, and Matt Gardner. 2019.
\newblock \href {https://doi.org/10.18653/v1/D19-1606} {{Q}uoref: A reading comprehension dataset with questions requiring coreferential reasoning}.
\newblock In \emph{Proceedings of the 2019 Conference on Empirical Methods in Natural Language Processing and the 9th International Joint Conference on Natural Language Processing (EMNLP-IJCNLP)}, pages 5925--5932, Hong Kong, China. Association for Computational Linguistics.

\bibitem[{Devlin et~al.(2019)Devlin, Chang, Lee, and Toutanova}]{devlin-etal-2019-bert}
Jacob Devlin, Ming-Wei Chang, Kenton Lee, and Kristina Toutanova. 2019.
\newblock \href {https://doi.org/10.18653/v1/N19-1423} {{BERT}: Pre-training of deep bidirectional transformers for language understanding}.
\newblock In \emph{Proceedings of the 2019 Conference of the North {A}merican Chapter of the Association for Computational Linguistics: Human Language Technologies, Volume 1 (Long and Short Papers)}, pages 4171--4186, Minneapolis, Minnesota. Association for Computational Linguistics.

\bibitem[{Gu et~al.(2018)Gu, Wang, Cho, and Li}]{gu2018search}
Jiatao Gu, Yong Wang, Kyunghyun Cho, and Victor~OK Li. 2018.
\newblock Search engine guided neural machine translation.
\newblock In \emph{Proceedings of the AAAI Conference on Artificial Intelligence}, volume~32.

\bibitem[{He et~al.(2021)He, Huang, Cui, Li, and Liu}]{he-etal-2021-fast}
Qiuxiang He, Guoping Huang, Qu~Cui, Li~Li, and Lemao Liu. 2021.
\newblock \href {https://doi.org/10.18653/v1/2021.acl-long.246} {Fast and accurate neural machine translation with translation memory}.
\newblock In \emph{Proceedings of the 59th Annual Meeting of the Association for Computational Linguistics and the 11th International Joint Conference on Natural Language Processing (Volume 1: Long Papers)}, pages 3170--3180, Online. Association for Computational Linguistics.

\bibitem[{Honovich et~al.(2021)Honovich, Choshen, Aharoni, Neeman, Szpektor, and Abend}]{honovich2021q}
Or~Honovich, Leshem Choshen, Roee Aharoni, Ella Neeman, Idan Szpektor, and Omri Abend. 2021.
\newblock ${Q^2}$: Evaluating factual consistency in knowledge-grounded dialogues via question generation and question answering.
\newblock \emph{arXiv preprint arXiv:2104.08202}.

\bibitem[{Hu et~al.(2021)Hu, Shen, Wallis, Allen-Zhu, Li, Wang, Wang, and Chen}]{hu2021lora}
Edward~J Hu, Yelong Shen, Phillip Wallis, Zeyuan Allen-Zhu, Yuanzhi Li, Shean Wang, Lu~Wang, and Weizhu Chen. 2021.
\newblock Lora: Low-rank adaptation of large language models.
\newblock \emph{arXiv preprint arXiv:2106.09685}.

\bibitem[{Hu et~al.(2019)Hu, Peng, Huang, and Li}]{hu-etal-2019-multi}
Minghao Hu, Yuxing Peng, Zhen Huang, and Dongsheng Li. 2019.
\newblock \href {https://doi.org/10.18653/v1/D19-1170} {A multi-type multi-span network for reading comprehension that requires discrete reasoning}.
\newblock In \emph{Proceedings of the 2019 Conference on Empirical Methods in Natural Language Processing and the 9th International Joint Conference on Natural Language Processing (EMNLP-IJCNLP)}, pages 1596--1606, Hong Kong, China. Association for Computational Linguistics.

\bibitem[{Huang et~al.(2015)Huang, Xu, and Yu}]{huang2015bidirectional}
Zhiheng Huang, Wei Xu, and Kai Yu. 2015.
\newblock Bidirectional lstm-crf models for sequence tagging. arxiv 2015.
\newblock \emph{arXiv preprint arXiv:1508.01991}.

\bibitem[{Lan et~al.(2019)Lan, Chen, Goodman, Gimpel, Sharma, and Soricut}]{lan2019albert}
Zhenzhong Lan, Mingda Chen, Sebastian Goodman, Kevin Gimpel, Piyush Sharma, and Radu Soricut. 2019.
\newblock Albert: A lite bert for self-supervised learning of language representations.
\newblock \emph{arXiv preprint arXiv:1909.11942}.

\bibitem[{Lee et~al.(2023)Lee, Kim, and Kang}]{lee2023liquid}
Seongyun Lee, Hyunjae Kim, and Jaewoo Kang. 2023.
\newblock Liquid: A framework for list question answering dataset generation.
\newblock \emph{arXiv preprint arXiv:2302.01691}.

\bibitem[{Li et~al.(2021)Li, Bi, Yan, Wang, and Huang}]{li-etal-2021-addressing-semantic}
Chenliang Li, Bin Bi, Ming Yan, Wei Wang, and Songfang Huang. 2021.
\newblock \href {https://doi.org/10.18653/v1/2021.acl-short.118} {Addressing semantic drift in generative question answering with auxiliary extraction}.
\newblock In \emph{Proceedings of the 59th Annual Meeting of the Association for Computational Linguistics and the 11th International Joint Conference on Natural Language Processing (Volume 2: Short Papers)}, pages 942--947, Online. Association for Computational Linguistics.

\bibitem[{Li et~al.(2022{\natexlab{a}})Li, Tomko, Vasardani, and Baldwin}]{li-etal-2022-multispanqa}
Haonan Li, Martin Tomko, Maria Vasardani, and Timothy Baldwin. 2022{\natexlab{a}}.
\newblock \href {https://doi.org/10.18653/v1/2022.naacl-main.90} {{M}ulti{S}pan{QA}: A dataset for multi-span question answering}.
\newblock In \emph{Proceedings of the 2022 Conference of the North American Chapter of the Association for Computational Linguistics: Human Language Technologies}, pages 1250--1260, Seattle, United States. Association for Computational Linguistics.

\bibitem[{Li et~al.(2022{\natexlab{b}})Li, Su, Cai, Wang, and Liu}]{li2022survey}
Huayang Li, Yixuan Su, Deng Cai, Yan Wang, and Lemao Liu. 2022{\natexlab{b}}.
\newblock A survey on retrieval-augmented text generation.
\newblock \emph{arXiv preprint arXiv:2202.01110}.

\bibitem[{Liu et~al.(2023{\natexlab{a}})Liu, Ning, Teng, Liu, Zhou, and Zhang}]{liu2023evaluating}
Hanmeng Liu, Ruoxi Ning, Zhiyang Teng, Jian Liu, Qiji Zhou, and Yue Zhang. 2023{\natexlab{a}}.
\newblock Evaluating the logical reasoning ability of chatgpt and gpt-4.
\newblock \emph{arXiv preprint arXiv:2304.03439}.

\bibitem[{Liu et~al.(2023{\natexlab{b}})Liu, Cho, Freedman, Ma, and May}]{liu-etal-2023-recap}
Shuai Liu, Hyundong Cho, Marjorie Freedman, Xuezhe Ma, and Jonathan May. 2023{\natexlab{b}}.
\newblock \href {https://doi.org/10.18653/v1/2023.acl-long.468} {{RECAP}: Retrieval-enhanced context-aware prefix encoder for personalized dialogue response generation}.
\newblock In \emph{Proceedings of the 61st Annual Meeting of the Association for Computational Linguistics (Volume 1: Long Papers)}, pages 8404--8419, Toronto, Canada. Association for Computational Linguistics.

\bibitem[{Liu et~al.(2023{\natexlab{c}})Liu, Yu, He, Zhang, Wei, Sun, and Tu}]{liu-etal-2023-system}
Xiao Liu, Junfeng Yu, Yibo He, Lujun Zhang, Kaiyichen Wei, Hongbo Sun, and Gang Tu. 2023{\natexlab{c}}.
\newblock \href {https://aclanthology.org/2023.ccl-3.34} {System report for {CCL}23-eval task 9: {HUST}1037 explore proper prompt strategy for {LLM} in {MRC} task}.
\newblock In \emph{Proceedings of the 22nd Chinese National Conference on Computational Linguistics (Volume 3: Evaluations)}, pages 310--319, Harbin, China. Chinese Information Processing Society of China.

\bibitem[{Nguyen et~al.(2016)Nguyen, Rosenberg, Song, Gao, Tiwary, Majumder, and Deng}]{nguyen2016ms}
Tri Nguyen, Mir Rosenberg, Xia Song, Jianfeng Gao, Saurabh Tiwary, Rangan Majumder, and Li~Deng. 2016.
\newblock Ms marco: A human generated machine reading comprehension dataset.
\newblock \emph{choice}, 2640:660.

\bibitem[{Ohsugi et~al.(2019)Ohsugi, Saito, Nishida, Asano, and Tomita}]{ohsugi-etal-2019-simple}
Yasuhito Ohsugi, Itsumi Saito, Kyosuke Nishida, Hisako Asano, and Junji Tomita. 2019.
\newblock \href {https://doi.org/10.18653/v1/W19-4102} {A simple but effective method to incorporate multi-turn context with {BERT} for conversational machine comprehension}.
\newblock In \emph{Proceedings of the First Workshop on NLP for Conversational AI}, pages 11--17, Florence, Italy. Association for Computational Linguistics.

\bibitem[{Rajpurkar et~al.(2016)Rajpurkar, Zhang, Lopyrev, and Liang}]{rajpurkar-etal-2016-squad}
Pranav Rajpurkar, Jian Zhang, Konstantin Lopyrev, and Percy Liang. 2016.
\newblock \href {https://doi.org/10.18653/v1/D16-1264} {{SQ}u{AD}: 100,000+ questions for machine comprehension of text}.
\newblock In \emph{Proceedings of the 2016 Conference on Empirical Methods in Natural Language Processing}, pages 2383--2392, Austin, Texas. Association for Computational Linguistics.

\bibitem[{Saha and Srihari(2023)}]{saha-srihari-2023-argu}
Sougata Saha and Rohini Srihari. 2023.
\newblock \href {https://doi.org/10.18653/v1/2023.acl-long.466} {{A}rg{U}: A controllable factual argument generator}.
\newblock In \emph{Proceedings of the 61st Annual Meeting of the Association for Computational Linguistics (Volume 1: Long Papers)}, pages 8373--8388, Toronto, Canada. Association for Computational Linguistics.

\bibitem[{Segal et~al.(2020)Segal, Efrat, Shoham, Globerson, and Berant}]{segal-etal-2020-simple}
Elad Segal, Avia Efrat, Mor Shoham, Amir Globerson, and Jonathan Berant. 2020.
\newblock \href {https://doi.org/10.18653/v1/2020.emnlp-main.248} {A simple and effective model for answering multi-span questions}.
\newblock In \emph{Proceedings of the 2020 Conference on Empirical Methods in Natural Language Processing (EMNLP)}, pages 3074--3080, Online. Association for Computational Linguistics.

\bibitem[{Su et~al.(2022)Su, Li, Zhang, Shang, Jiang, Liu, and Fung}]{su-etal-2022-read}
Dan Su, Xiaoguang Li, Jindi Zhang, Lifeng Shang, Xin Jiang, Qun Liu, and Pascale Fung. 2022.
\newblock \href {https://doi.org/10.18653/v1/2022.findings-acl.61} {Read before generate! faithful long form question answering with machine reading}.
\newblock In \emph{Findings of the Association for Computational Linguistics: ACL 2022}, pages 744--756, Dublin, Ireland. Association for Computational Linguistics.

\bibitem[{Su et~al.(2021)Su, Wang, Cai, Baker, Korhonen, and Collier}]{su2021prototype}
Yixuan Su, Yan Wang, Deng Cai, Simon Baker, Anna Korhonen, and Nigel Collier. 2021.
\newblock Prototype-to-style: Dialogue generation with style-aware editing on retrieval memory.
\newblock \emph{IEEE/ACM Transactions on Audio, Speech, and Language Processing}, 29:2152--2161.

\bibitem[{Taori et~al.(2023)Taori, Gulrajani, Zhang, Dubois, Li, Guestrin, Liang, and Hashimoto}]{alpaca}
Rohan Taori, Ishaan Gulrajani, Tianyi Zhang, Yann Dubois, Xuechen Li, Carlos Guestrin, Percy Liang, and Tatsunori~B. Hashimoto. 2023.
\newblock Stanford alpaca: An instruction-following llama model.
\newblock \url{https://github.com/tatsu-lab/stanford_alpaca}.

\bibitem[{Touvron et~al.(2023)Touvron, Martin, Stone, Albert, Almahairi, Babaei, Bashlykov, Batra, Bhargava, Bhosale et~al.}]{touvron2023llama}
Hugo Touvron, Louis Martin, Kevin Stone, Peter Albert, Amjad Almahairi, Yasmine Babaei, Nikolay Bashlykov, Soumya Batra, Prajjwal Bhargava, Shruti Bhosale, et~al. 2023.
\newblock Llama 2: Open foundation and fine-tuned chat models.
\newblock \emph{arXiv preprint arXiv:2307.09288}.

\bibitem[{Wang et~al.(2018)Wang, Yan, and Wu}]{wang-etal-2018-multi-granularity}
Wei Wang, Ming Yan, and Chen Wu. 2018.
\newblock \href {https://doi.org/10.18653/v1/P18-1158} {Multi-granularity hierarchical attention fusion networks for reading comprehension and question answering}.
\newblock In \emph{Proceedings of the 56th Annual Meeting of the Association for Computational Linguistics (Volume 1: Long Papers)}, pages 1705--1714, Melbourne, Australia. Association for Computational Linguistics.

\bibitem[{Weston et~al.(2018)Weston, Dinan, and Miller}]{weston-etal-2018-retrieve}
Jason Weston, Emily Dinan, and Alexander Miller. 2018.
\newblock \href {https://doi.org/10.18653/v1/W18-5713} {Retrieve and refine: Improved sequence generation models for dialogue}.
\newblock In \emph{Proceedings of the 2018 {EMNLP} Workshop {SCAI}: The 2nd International Workshop on Search-Oriented Conversational {AI}}, pages 87--92, Brussels, Belgium. Association for Computational Linguistics.

\bibitem[{Wu et~al.(2021)Wu, Galley, Brockett, Zhang, Gao, Quirk, Koncel-Kedziorski, Gao, Hajishirzi, Ostendorf et~al.}]{wu2021controllable}
Zeqiu Wu, Michel Galley, Chris Brockett, Yizhe Zhang, Xiang Gao, Chris Quirk, Rik Koncel-Kedziorski, Jianfeng Gao, Hannaneh Hajishirzi, Mari Ostendorf, et~al. 2021.
\newblock A controllable model of grounded response generation.
\newblock In \emph{Proceedings of the AAAI Conference on Artificial Intelligence}, volume~35, pages 14085--14093.

\bibitem[{Xiao et~al.(2021)Xiao, Pang, Lan, Wang, Shen, and Cheng}]{xiao-etal-2021-transductive}
Fei Xiao, Liang Pang, Yanyan Lan, Yan Wang, Huawei Shen, and Xueqi Cheng. 2021.
\newblock \href {https://doi.org/10.18653/v1/2021.emnlp-main.195} {Transductive learning for unsupervised text style transfer}.
\newblock In \emph{Proceedings of the 2021 Conference on Empirical Methods in Natural Language Processing}, pages 2510--2521, Online and Punta Cana, Dominican Republic. Association for Computational Linguistics.

\bibitem[{Yan et~al.(2019)Yan, Xia, Wu, Bi, Zhao, Zhang, Si, Wang, Wang, and Chen}]{yan2019deep}
Ming Yan, Jiangnan Xia, Chen Wu, Bin Bi, Zhongzhou Zhao, Ji~Zhang, Luo Si, Rui Wang, Wei Wang, and Haiqing Chen. 2019.
\newblock A deep cascade model for multi-document reading comprehension.
\newblock In \emph{Proceedings of the AAAI conference on artificial intelligence}, volume~33, pages 7354--7361.

\bibitem[{Yang et~al.(2020)Yang, Zhang, and Zhao}]{yang2020multi}
Junjie Yang, Zhuosheng Zhang, and Hai Zhao. 2020.
\newblock Multi-span style extraction for generative reading comprehension.
\newblock \emph{arXiv preprint arXiv:2009.07382}.

\bibitem[{Ye et~al.(2020)Ye, Lin, Du, Liu, Li, Sun, and Liu}]{ye-etal-2020-coreferential}
Deming Ye, Yankai Lin, Jiaju Du, Zhenghao Liu, Peng Li, Maosong Sun, and Zhiyuan Liu. 2020.
\newblock \href {https://doi.org/10.18653/v1/2020.emnlp-main.582} {{C}oreferential {R}easoning {L}earning for {L}anguage {R}epresentation}.
\newblock In \emph{Proceedings of the 2020 Conference on Empirical Methods in Natural Language Processing (EMNLP)}, pages 7170--7186, Online. Association for Computational Linguistics.

\bibitem[{Zhang et~al.(2023)Zhang, Lin, Liu, Lai, Feng, and Zhao}]{zhang2023many}
Chen Zhang, Jiuheng Lin, Xiao Liu, Yuxuan Lai, Yansong Feng, and Dongyan Zhao. 2023.
\newblock How many answers should i give? an empirical study of multi-answer reading comprehension.
\newblock \emph{arXiv preprint arXiv:2306.00435}.

\bibitem[{Zhu et~al.(2023)Zhu, Xu, Huang, Kong, and Chen}]{zhu-etal-2023-ink}
Wenhao Zhu, Jingjing Xu, Shujian Huang, Lingpeng Kong, and Jiajun Chen. 2023.
\newblock \href {https://doi.org/10.18653/v1/2023.acl-long.888} {{INK}: Injecting k{NN} knowledge in nearest neighbor machine translation}.
\newblock In \emph{Proceedings of the 61st Annual Meeting of the Association for Computational Linguistics (Volume 1: Long Papers)}, pages 15948--15959, Toronto, Canada. Association for Computational Linguistics.

\end{thebibliography}

\clearpage
\appendix

\section{Detailed Experiment Setup and Results}
\label{sec:appendix}

\subsection{Dataset Leaderboard}
\label{subsec:leaderboard}

Below are the official leaderboards all the datasets we refer to:

\begin{table}[!htbp]
    \centering
    \small
    \begin{tabular}{lp{2in}}
    \hline\hline
    \textbf{SQuAD} & \url{https://rajpurkar.github.io/SQuAD-explorer/} \\\hline
    \textbf{MultiSpanQA} & \url{https://multi-span.github.io/} \\\hline
    \textbf{Quoref} & \url{https://leaderboard.allenai.org/quoref/submissions/public} \\\hline\hline
    \end{tabular}
    \caption{Dataset official leaderboards.}
    \label{tab:leaderboards}
\end{table}

\subsection{Hyper-Parameter Selection}
\label{subsec:hp_selection}
In this section, we outline the process for selecting the hyper-parameter $\beta$ and detail our approach to LoRA fine-tuning.

For selecting $\beta$, we use a grid search method, exploring values from 0.5 to 2 in increments of 0.1, on 30\% of the MultiSpanQA training dataset. This process leads to the determination that $\beta = 1$ empirically yield the best performance, hence it is selected for use in our experiments.

To select the learning rate $lr$, we conduct a grid search, testing values from $\{1e-5, 5e-5, 1e-4, 5e-4, 1e-3\}$ on 30\% of the MultiSpanQA training dataset. Empirically, the value $1e-4$ demonstrates the best performance and is therefore chosen for our experiments. This selection is in agreement with the default $lr$ value used in Meta's official Llama 2 fine-tuning recipe\footnote{\href{https://github.com/facebookresearch/llama-recipes/blob/main/src/llama_recipes/configs/training.py}{Link to the fine-tuning configuration of Meta's official Llama 2 recipe.}}.

In the case of LoRA fine-tuning, we follow the established methodology as outlined by \citet{hu2021lora}. This involves applying LoRA to Llama 2 and the pre-trained Alpaca models by freezing their pre-trained weights and integrating trainable rank decomposition matrices at every layer of their Transformer structures, aimed at reducing the number of trainable parameters to enhance computational efficiency. We implement this using the PEFT package\footnote{\href{https://github.com/huggingface/peft}{Link to the Hugging Face PEFT implementation.}}. The fine-tuning hyper-parameters for LoRA are set according to the default settings specified in Meta's official Llama 2 fine-tuning recipe\footnote{\href{src/llama_recipes/configs/peft.py}{Link to the LoRA hyper-parameter configuration of Meta's official Llama 2 recipe.}}, which include a rank $r = 8$, $\alpha = 32$, and a dropout rate of $0.05$.

\subsection{Full Experiment Results}
\label{subsec:full_results}
In addition to the highlighted results presented in Section \ref{sec:experiments}, we also compare the fine-tuned PLMs to their corresponding base PLMs in zero-shot settings. The results, presented in Table \ref{tab:plm_qase}, show that fine-tuning with \textit{QASE} improves performance across all datasets. Specifically, on the SQuAD dataset, models using \textit{QASE} perform up to 5.6 times better in exact match and 3.0 times better in F1 score compared to the original models. On the MultiSpanQA dataset, the exact match improves by up to 124.4 times, and F1 score by up to 3.4 times. Similarly, on the Quoref dataset, the exact match improves by up to 38.4 times, and F1 score by up to 11.2 times with \textit{QASE}.

\begin{table*}[!htbp]
    \centering
    \small
    \begin{tabular}{l||cc|cc|cc}
    \hline\hline
    \textbf{}          & \multicolumn{2}{c|}{\textbf{MultiSpanQA}} & \multicolumn{2}{c|}{\textbf{SQuAD}} & \multicolumn{2}{c}{\textbf{Quoref}} \\
    \textbf{}          & \textbf{EM F1}    & \textbf{Overlap F1}  & \textbf{EM}      & \textbf{F1}     & \textbf{EM}      & \textbf{F1}      \\ \hline\hline
    Llama2        & 7.354  & 34.031 & 13.443 & 28.931 & 5.02  & 28.91 \\
    Llama2$_{FT}$ & 50.934 & 68.140 & 36.679 & 47.055 & 45.52 & 52.09 \\
    Llama2$_{QASE}$       & \textbf{51.748}   & \textbf{70.389}      & \textbf{37.219}  & \textbf{47.686} & \textbf{54.28}   & \textbf{60.44}   \\ \hline
    Alpaca        & 15.201 & 42.759 & 18.259 & 33.871 & -     & -     \\
    Alpaca$_{FT}$ & \textbf{52.730} & 69.099 & 27.881 & 43.950 & - & - \\
    Alpaca$_{QASE}$        & 52.196   & \textbf{70.008}      & \textbf{37.313}  & \textbf{47.622} & -                & -                \\ \hline
    Flan-T5-Small & 0.475  & 22.539 & 13.878 & 28.710 & 1.58  & 5.96  \\
    Flan-T5-Small$_{FT}$ & \textbf{59.128} & 76.494 & 77.332 & 85.513 & 58.21 & 63.30 \\
    Flan-T5-Small$_{QASE}$ & 59.080   & \textbf{77.103}      & \textbf{77.663}  & \textbf{85.901} & \textbf{60.70}   & \textbf{66.88}   \\ \hline
    Flan-T5-Base  & 4.113  & 37.694 & 37.596 & 51.747 & 27.08 & 34.38 \\
    Flan-T5-Base$_{FT}$ & 64.659 & 81.408 & 82.090 & 89.558 & 72.77 & 80.90 \\
    Flan-T5-Base$_{QASE}$  & \textbf{64.874}   & \textbf{81.498}      & \textbf{82.204}  & \textbf{90.240} & \textbf{75.17}   & \textbf{81.18}   \\ \hline
    Flan-T5-Large & 13.907 & 51.501 & 16.149 & 37.691 & 15.96 & 24.10 \\
    Flan-T5-Large$_{FT}$ & \textbf{67.408} & 83.094 & 83.159 & 90.712 & 75.17 & 80.49 \\
    Flan-T5-Large$_{QASE}$ & 66.918   & \textbf{84.221}      & \textbf{84.125}  & \textbf{91.701} & \textbf{76.19}   & \textbf{82.13}   \\ \hline\hline
    \end{tabular}
    \caption{Performance of zero-shot PLMs and fined-tuned PLMs with and without \textit{QASE}.}
    \label{tab:plm_qase}
\end{table*}

\subsection{Instruction Templates and Model Prompts}
\label{subsec:prompts}
Table \ref{tab:prompts} provides the instruction and prompt templates used for fine-tuning the PLMs and for zero-shot querying of PLMs and GPT variants across both single- and multi-span answer datasets.

\begin{table*}
    \centering
    \small
    \begin{tabular}{p{1.8in}|p{4.2in}}
        \hline\hline
        \textbf{Fine-tuning} PLMs & Instruction: Using the provided context, answer the question with exact phrases and avoid explanations.\newline - - - \newline Context: \{context\}\newline - - - \newline Question: \{question\}\newline - - -\newline Answer: \\\hline
        \textbf{Zero-shot} prompting PLMs and GPT variants on \textbf{single-span} answer dataset, SQuAD & Instruction: Using the provided context, answer the question with exact phrases and avoid explanations.\newline - - - \newline Context: \{context\}\newline - - - \newline Question: \{question\}\newline - - -\newline Answer: \\\hline
        \textbf{Zero-shot} prompting PLMs and GPT variants on \textbf{multi-span} answer datasets, MultiSpanQA and Quoref & Instruction: Using the provided context, answer the question with exact phrases and avoid explanations. Format the response as follows: ["answer1", "answer2", ...].\newline - - - \newline Context: \{context\}\newline - - - \newline Question: \{question\}\newline - - -\newline Answer: \\\hline\hline
    \end{tabular}
    \caption{Templates for fine-tuning instructions and zero-shot query prompts}
    \label{tab:prompts}
\end{table*}

\subsection{Ablation Studies Details}
\label{subsec:ablation_details}
\begin{figure}[!htbp]
  \centering
  \includegraphics[width=0.5\textwidth]{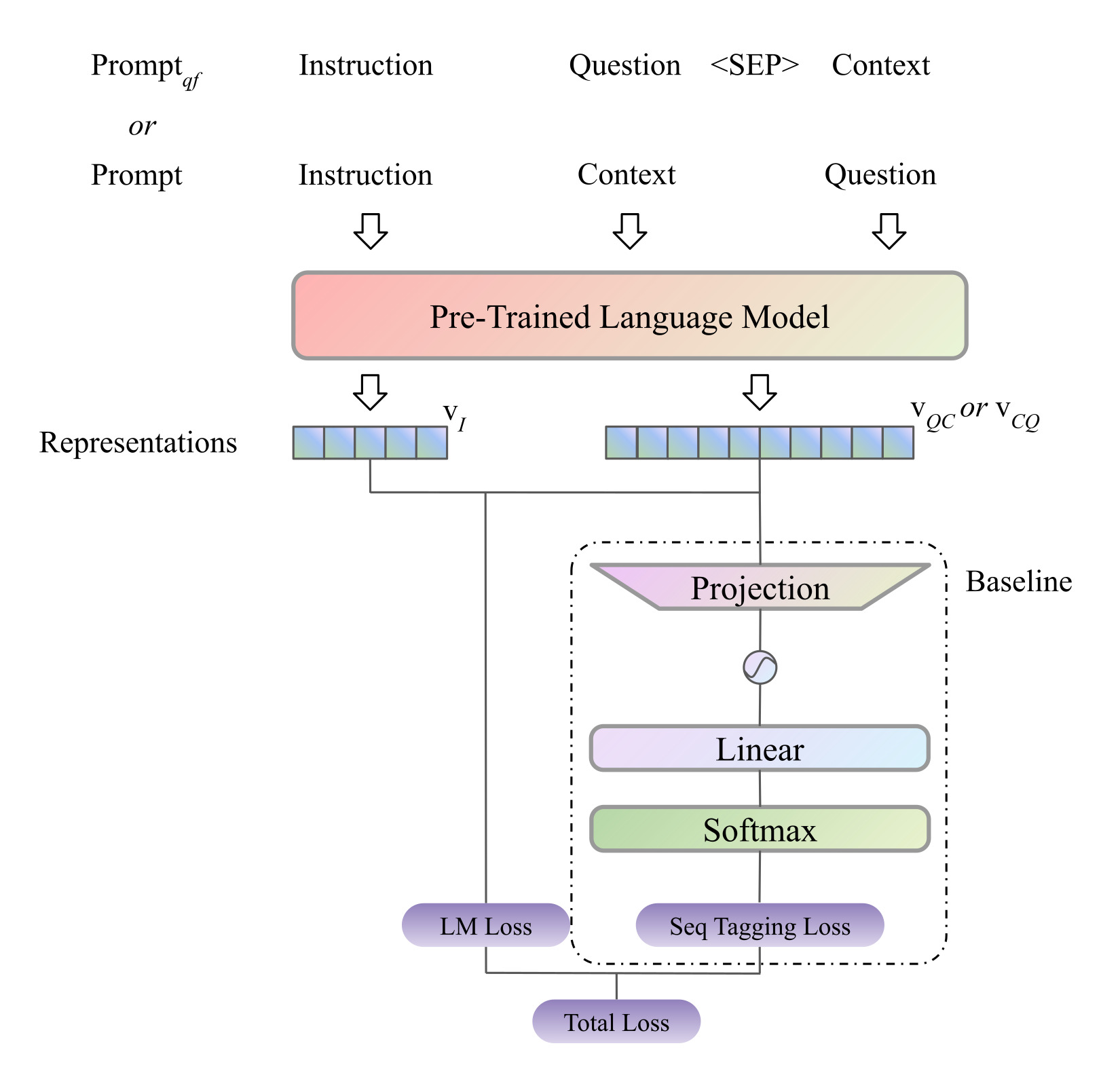}
  \caption{Baseline-embedded model architecture}
  \label{fig:baseline}
\end{figure}

Figure \ref{fig:baseline} depicts the architecture of the model we use for the ablation studies, with a baseline span extraction module. The baseline span extraction module omits the \textit{MHA} component, typifying a standard architecture for fine-tuning pre-trained encoders for downstream sequence tagging tasks. The baseline-embedded Flan-T5-Large models are fine-tuned with the same configurations as Flan-T5-Large$_{QASE}$ including learning rate, weight decay, batch size, epoch number, and GPU type.

We experiment with 2 prompting strategies for ablation studies:
\begin{itemize}
    \item \textbf{Context-first prompting:} The default prompting strategy we utilize for fine-tuning PLMs, both with and without \textit{QASE}. In this setting, the prompt is ordered as "<instruction tokens> <context tokens> <question tokens>".
    \item \textbf{Question-first prompting (\textit{qf}):} Following BERT's standard fine-tuning procedures. In this setting, the prompt is ordered as "<instruction tokens> <question tokens> <SEP> <context tokens>". <SEP> is a special separator token.
\end{itemize}

\newpage
\subsection{Factual Consistency Case Studies}
\label{subsec:case_studies}
In Section \ref{sec:experiments}, we demonstrate that the Flan-T5-Large model, when fine-tuned with \textit{QASE}, produces answers with greater factual accuracy in relation to the context compared to its counterpart fine-tuned without \textit{QASE}. Specifically, we observe a 1.0\% improvement in the $Q^2$ score on the SQuAD dataset and a significant 16.0\% increase on MultiSpanQA. This section includes examples to further illustrate \textit{QASE}'s effectiveness.

Table \ref{tab:showcase_attention} showcases that Flan-T5-Large$_{QASE}$ more accurately identifies the key focus of the question and locates the pertinent factual information within the context, with the aid of the \textit{QASE} module. For instance, in \textbf{Sample 1}, Flan-T5-Large$_{QASE}$ correctly interprets the question as seeking the age difference between Newton and Manning, rather than the age of either individual, and accordingly provides the accurate answer. In contrast, Flan-T5-Large$_{FT}$ mistakenly provides Newton's age as the answer. Similarly, in \textbf{Sample 2}, Flan-T5-Large$_{QASE}$ accurately discerns that the question pertains to Thoreau's claim regarding the majority, generating in the correct answer, whereas Flan-T5-Large$_{FT}$ misguidedly responds with Thoreau's political philosophy.

\begin{table}[!htbp]
    \centering
    \small
    \begin{tabular}{c||p{1.5in}}
        \hline\hline
        \multicolumn{2}{c}{\textbf{Sample 1}} \\
        \multicolumn{2}{p{2.8in}}{\textbf{Context:} This was the first Super Bowl to feature a quarterback on both teams who was the \#1 pick in their draft classes. Manning was the \#1 selection of the 1998 NFL draft, while Newton was picked first in 2011. The matchup also pits the top two picks of the 2011 draft against each other: Newton for Carolina and Von Miller for Denver. Manning and Newton also set the record for the largest \hl{age difference} between opposing Super Bowl quarterbacks at 13 years and 48 days (Manning was 39, Newton was 26).} \\
        \multicolumn{2}{p{2.8in}}{\textbf{Question:} What was the \hl{age difference} between Newton and Manning in Super Bowl 50?} \\\hline
        \multicolumn{2}{p{2.8in}}{\textcolor{green!75!blue}{\textbf{Gold Answer:} 13 years and 48 days}} \\\hline\hline
        \begin{tabular}[c]{@{}c@{}}\textbf{Flan-T5-Large}$\boldsymbol{_{QASE}}$\\\textbf{Generation}\end{tabular} & \textcolor{green!75!blue}{13 years and 48 days} \\\hline
        \begin{tabular}[c]{@{}c@{}}Flan-T5-Large$_{FT}$\\Generation\end{tabular} & \textcolor{red!90!green}{26} \\\hline\hline
        \multicolumn{2}{c}{} \\\hline\hline

        \multicolumn{2}{c}{\textbf{Sample 2}} \\
        \multicolumn{2}{p{2.8in}}{\textbf{Context:} However, this definition is disputed by Thoreau's political philosophy, which contrasts the conscience with the collective. The individual is the ultimate arbiter of right and wrong. Beyond this, since only individuals act, only they can commit injustices. When the government knocks on the door, it is an individual in the guise of a postman or tax collector whose hand meets the wood. Before Thoreau's imprisonment, when a perplexed tax collector openly pondered how to deal with his refusal to pay, Thoreau had advised, "Resign." If a man chose to be an agent of injustice, then Thoreau insisted on confronting him with the reality that he was making a choice. But if the government is "the voice of the people," as often claimed, shouldn't that voice be heeded? Thoreau acknowledges that the government may represent the will of the majority but it might also merely reflect the desires of elite politicians. Even a good government is "liable to be abused and perverted before the people can act through it." Furthermore, even if a government did express the voice of the people, this fact would not obligate the obedience of individuals who dissent. \hl{The majority may be powerful but it is not necessarily right.} What, then, is the appropriate relationship between the individual and the government?} \\
        \multicolumn{2}{p{2.8in}}{\textbf{Question:} What did Thoreau claim about \hl{the majority?}} \\\hline
        \multicolumn{2}{p{2.8in}}{\textcolor{green!75!blue}{\textbf{Gold Answer:} not necessarily right}} \\\hline\hline
        \begin{tabular}[c]{@{}c@{}}\textbf{Flan-T5-Large}$\boldsymbol{_{QASE}}$\\\textbf{Generation}\end{tabular} & it is \textcolor{green!75!blue}{not necessarily right} \\\hline
        \begin{tabular}[c]{@{}c@{}}Flan-T5-Large$_{FT}$\\Generation\end{tabular} & \textcolor{red!90!green}{conscience vs. the collective} \\\hline\hline
    \end{tabular}
    \caption{Comparisons of model attention alignment with question key aspects and relevant factual context between Flan-T5-Large$_{QASE}$ and Flan-T5-Large$_{FT}$.}
    \label{tab:showcase_attention}
\end{table}

Flan-T5-Large$_{QASE}$ also shows a notable improvement in comprehending complex, lengthy sentences and synthesizing answers from information that is sparsely distributed across multiple spans requiring logical processing. This capability is particularly valuable when the answer to a question does not directly stem from a single phrase. Table \ref{tab:showcase_multispan} provides examples of such instances. In \textbf{Sample 3}, the model needs to recognize that ESPN Deportes is the exclusive broadcaster in Spanish and that CBS, although mentioned, does not offer Spanish-language broadcasting. Combining these facts leads to the correct answer, that ESPN Deportes is the network that broadcast the game in Spanish. Flan-T5-Large$_{QASE}$ accurately generates this answer, whereas Flan-T5-Large$_{FT}$ incorrectly answers with "CBS", likely due to confusion caused by the complex sentence structures and dispersed information. Similarly, in \textbf{Sample 4}, Flan-T5-Large$_{QASE}$ correctly identifies the question as seeking the name of the force related to a potential field between two locations. It successfully locates the relevant long sentence, deconstructs, and comprehends it to produce the correct answer, in contrast to Flan-T5-Large$_{FT}$, which incorrectly selects the first phrase mentioning "force". In \textbf{Sample 5}, the question asks for the class most commonly not ascribed to the graph isomorphism problem. The model needs to deduce from the context that "it is widely believed that the polynomial hierarchy does not collapse to any finite level", implying "graph isomorphism is not NP-complete". Once again, Flan-T5-Large$_{QASE}$ arrives at the correct conclusion, while Flan-T5-Large$_{FT}$ does not.

\begin{table}[!htbp]
    \centering
    \small
    \begin{tabular}{c||p{1.5in}}
        \hline\hline
        \multicolumn{2}{c}{\textbf{Sample 3}} \\
        \multicolumn{2}{p{2.8in}}{\textbf{Context:} On December 28, 2015, \hl{ESPN Deportes} announced that they had reached an agreement with CBS and the NFL to be \hl{the exclusive Spanish-language broadcaster} of the game, marking the third dedicated Spanish-language broadcast of the Super Bowl. Unlike NBC and Fox, \hl{CBS does not have a Spanish-language outlet of its own} that could broadcast the game (though per league policy, a separate Spanish play-by-play call was carried on CBS's second audio program channel for over-the-air viewers). The game was called by ESPN Deportes' Monday Night Football commentary crew of Alvaro Martin and Raul Allegre, and sideline reporter John Sutcliffe. ESPN Deportes broadcast pre-game and post-game coverage, while Martin, Allegre, and Sutcliffe contributed English-language reports for ESPN's SportsCenter and Mike \& Mike.} \\
        \multicolumn{2}{p{2.8in}}{\textbf{Question:} Which network broadcast the game \hl{in Spanish?}} \\\hline
        \multicolumn{2}{p{2.8in}}{\textcolor{green!75!blue}{\textbf{Gold Answer:} ESPN Deportes}} \\\hline\hline
        \begin{tabular}[c]{@{}c@{}}\textbf{Flan-T5-Large}$\boldsymbol{_{QASE}}$\\\textbf{Generation}\end{tabular} & \textcolor{green!75!blue}{ESPN Deportes} \\\hline
        \begin{tabular}[c]{@{}c@{}}Flan-T5-Large$_{FT}$\\Generation\end{tabular} & \textcolor{red!90!green}{CBS} \\\hline\hline
        \multicolumn{2}{c}{} \\\hline\hline
        
        \multicolumn{2}{c}{\textbf{Sample 4}} \\
        \multicolumn{2}{p{2.8in}}{\textbf{Context:} A conservative force that acts on a closed system has an associated mechanical work that allows energy to convert only between kinetic or potential forms. This means that for a closed system, the net mechanical energy is conserved whenever a conservative force acts on the system. \hl{The force}, therefore, is \hl{related directly to the difference in potential energy between two different locations} in space, and \hl{can be considered to be an artifact} of the potential field in the same way that the direction and amount of a flow of water can be considered to be an artifact of the contour map of the elevation of an area.} \\
        \multicolumn{2}{p{2.8in}}{\textbf{Question:} What is \hl{the force} called \hl{regarding a potential field between two locations}?} \\\hline
        \multicolumn{2}{p{2.8in}}{\textcolor{green!75!blue}{\textbf{Gold Answer:} an artifact}} \\\hline\hline
        \begin{tabular}[c]{@{}c@{}}\textbf{Flan-T5-Large}$\boldsymbol{_{QASE}}$\\\textbf{Generation}\end{tabular} & \textcolor{green!75!blue}{an artifact} \\\hline
        \begin{tabular}[c]{@{}c@{}}Flan-T5-Large$_{FT}$\\Generation\end{tabular} & \textcolor{red!90!green}{conservative force} \\\hline\hline

        \multicolumn{2}{c}{} \\\hline\hline
        
        \multicolumn{2}{c}{\textbf{Sample 5}} \\
        \multicolumn{2}{p{2.8in}}{\textbf{Context:} The graph isomorphism problem is the computational problem of determining whether two finite graphs are isomorphic. An important unsolved problem in complexity theory is whether the graph isomorphism problem is in P, NP-complete, or NP-intermediate. The answer is not known, but \hl{it is believed that the problem is at least not NP-complete.} If graph isomorphism is NP-complete, the polynomial time hierarchy collapses to its second level. Since \hl{it is widely believed that the polynomial hierarchy does not collapse to any finite level}, it is believed that \hl{graph isomorphism is not NP-complete.} The best algorithm for this problem, due to Laszlo Babai and Eugene Luks has run time $2O(\sqrt{n log(n)})$ for graphs with n vertices.} \\
        \multicolumn{2}{p{2.8in}}{\textbf{Question:} What class is \hl{most commonly not ascribed to the graph isomorphism problem} in spite of definitive determination?} \\\hline
        \multicolumn{2}{p{2.8in}}{\textcolor{green!75!blue}{\textbf{Gold Answer:} NP-complete}} \\\hline\hline
        \begin{tabular}[c]{@{}c@{}}\textbf{Flan-T5-Large}$\boldsymbol{_{QASE}}$\\\textbf{Generation}\end{tabular} & \textcolor{green!75!blue}{NP-complete} \\\hline
        \begin{tabular}[c]{@{}c@{}}Flan-T5-Large$_{FT}$\\Generation\end{tabular} & \textcolor{red!90!green}{NP-intermediate} \\\hline\hline
    \end{tabular}
    \caption{Comparison of Flan-T5-Large$_{QASE}$ and Flan-T5-Large$_{FT}$ in understanding complex sentence structures.}
    \label{tab:showcase_multispan}
\end{table}

While our primary evaluation focuses on the model's proficiency in deriving answers from provided contexts, we also note that \textit{QASE} enhances the model's capacity to leverage real-world knowledge acquired during its pre-training phase. This improvement is attributed to \textit{QASE}'s ability to better align the model's focus on parts of the context that are relevant to the questions asked. Table \ref{tab:showcase_knowledge} presents an example of this phenomenon. In \textbf{Sample 6}, when asked about the California venue considered for the Super Bowl, Flan-T5-Large$_{QASE}$ correctly associates the San Francisco Bay Area with California, thus producing the accurate answer. On the other hand, Flan-T5-Large$_{FT}$ erroneously identifies a stadium in Miami as the answer. This example illustrates how \textit{QASE} not only improves context-based answer generation but also the model's application of pre-existing real-world knowledge to the questions posed.

\begin{table}[!htbp]
    \centering
    \small
    \begin{tabular}{c||p{1.5in}}
        \hline\hline
        \multicolumn{2}{c}{\textbf{Sample 6}} \\
        \multicolumn{2}{p{2.8in}}{\textbf{Context:} The league eventually narrowed the bids to three sites: New Orleans' Mercedes-Benz Superdome, Miami's Sun Life Stadium, and the \hl{San Francisco Bay Area's} Levi's Stadium.} \\
        \multicolumn{2}{p{2.8in}}{\textbf{Question:} Which \hl{California} venue was one of three considered for Super Bowl 50?} \\\hline
        \multicolumn{2}{p{2.8in}}{\textcolor{green!75!blue}{\textbf{Gold Answer:} San Francisco Bay Area's Levi's Stadium}} \\\hline\hline
        \begin{tabular}[c]{@{}c@{}}\textbf{Flan-T5-Large}$\boldsymbol{_{QASE}}$\\\textbf{Generation}\end{tabular} & \textcolor{green!75!blue}{San Francisco Bay Area's Levi's Stadium} \\\hline
        \begin{tabular}[c]{@{}c@{}}Flan-T5-Large$_{FT}$\\Generation\end{tabular} & \textcolor{red!90!green}{Sun Life Stadium} \\\hline\hline
    \end{tabular}
    \caption{Comparison of Flan-T5-Large$_{QASE}$ and Flan-T5-Large$_{FT}$ in utilizing real-world knowledge.}
    \label{tab:showcase_knowledge}
\end{table}

\section{Extended Discussion on Model Performance}
In this section, we engage in a detailed discussion on the performance of the Flan-T5 family of models and Llama 2 in MRC tasks. Our aim is to gain insights into the reasons behind the modest zero-shot performance of these large PLMs on MRC tasks, despite their adeptness at handling other complex NLP tasks such as dialogue generation and summarization. Although a comprehensive analysis falls outside the scope of our current study, exploring these performance nuances can provide valuable perspectives on how to potentially enhance the effectiveness of these PLMs on similar tasks.

\subsection{Discussion on Flan-T5 Zero-Shot Performance}
We observe that the zero-shot performance of Flan-T5 models across all datasets, including SQuAD, remains low as shown in Table \ref{tab:plm_qase}, despite being instruct-tuned on the SQuAD dataset during the pre-training phase. This underperformance might stem from the fact that Flan-T5 models, although trained on the <SQuAD, Extractive QA> task, are also trained on a broad spectrum of 1,836 tasks, predominantly focusing on free-form generation, QA, and reasoning tasks \cite{chung2022scaling}. Consequently, these models are not finely optimized for extractive QA tasks like MRC, especially under metrics like exact match and F1, particularly for the smaller to larger variants under study. The larger XL and XXL variants may exhibit better performance in these tasks. Furthermore, as discussed in the previous sections, generative models, including Llama 2, Alpaca, and GPT variants, generally show limited effectiveness in MRC tasks in zero-shot settings, underscored by their poorer performance despite having significantly larger model parameters compared to the Flan-T5 variants we experiment with.

To ensure that our zero-shot experiment's prompts do not adversely affect Flan-T5's performance, we compare our prompt template, detailed in Table \ref{tab:prompts}, with those Google released for Flan-T5's instruct-tuning on the SQuAD v1 dataset\footnote{\href{https://github.com/google-research/FLAN/blob/main/flan/templates.py}{Link to Flan-T5 instruct-tuning prompt templates.}}. Our template, similar to Google's, differs mainly by including "with exact phrases and avoid explanations." This difference could potentially affect performance, yet our subsequent experiments demonstrate otherwise.

We conduct a series of experiments to assess the zero-shot performance of Flan-T5-Large on SQuAD, using Google released templates for Flan-T5 instruct-tuning. We select three templates of varying complexities, as listed in Table \ref{tab:squad_zeroshot_prompts}. Our results, detailed in Table \ref{tab:squad_zeroshot_prompts}, reveal that our template achieves the highest F1 score. This indicates the lower performance of zero-shot Flan-T5 on SQuAD and similar MRC datasets is expected, even with the original instruct-tuning templates. It supports our hypothesis that, although Flan-T5 is instruct-tuned on SQuAD, its primary strengths are in broader generative question answering and reasoning, rather than specific extractive QA tasks such as MRC, particularly when evaluated by exact match and F1 metrics.

\begin{table}[!htbp]
    \centering
    \small
    \begin{tabular}{p{1in}|cc}
        \hline\hline
         & \multicolumn{2}{c}{\textbf{SQuAD Performance}} \\
        \textbf{Prompt Template} & \textbf{EM} & \textbf{F1} \\\hline\hline
        \begin{tabular}[c]{@{}l@{}}Article: \{context\}\\Question: \{question\}\\Answer:\end{tabular} & 7.001 & 21.717 \\\hline
        \begin{tabular}[c]{@{}l@{}}Answer a question\\about this article.\\Article: \{context\}\\Question: \{question\}\\Answer:\end{tabular} & 15.875 & 33.375 \\\hline
        \begin{tabular}[c]{@{}l@{}}Here is a question\\about this article:\\Article: \{context\}\\What is the answer\\to this question:\\Question: \{question\}\\Answer:\end{tabular} & \textbf{16.764} & 35.304 \\\hline\hline
        \begin{tabular}[c]{@{}l@{}}Our Template\\See Table \ref{tab:prompts}\end{tabular} & 16.149 & \textbf{37.691} \\\hline
    \end{tabular}
    \caption{Flan-T5-Large zero-shot performance on SQuAD with different prompt templates.}
    \label{tab:squad_zeroshot_prompts}
\end{table}

\subsection{Discussion on Llama 2 Performance}

We observe that models based on Llama 2 and Alpaca generally underperform compared to those based on Flan-T5, in both zero-shot and fine-tuned scenarios, with or without \textit{QASE}. This section delves into a detailed discussion of the potential reasons behind this trend.

Firstly, the discrepancy in performance may stem from the inherent structural differences between decoder-only models (Llama 2 and Alpaca) and encoder-decoder models (Flan-T5). Encoder-decoder models are better equipped for tasks that require extensive input processing, such as MRC, making them more apt for these tasks than decoder-only models, which are typically more suited to open-ended QA scenarios. This fundamental distinction partially accounts for Flan-T5's superior performance in context-based question answering across both zero-shot and fine-tuned settings.

Additionally, the difference in the number of trainable parameters during fine-tuning might contribute to the observed performance gap. Table \ref{tab:params} indicates that fine-tuning Llama 2 and Alpaca with LoRA leads to a significantly lower count of trainable parameters (4.2M) compared to even the smallest Flan-T5 model (77.0M). This disparity in trainable parameters is a crucial factor in explaining why fine-tuned Flan-T5 models, irrespective of the use of QASE, outperform Llama 2 and Alpaca models.

While we address these factors, conducting a comprehensive comparison and analysis of different generative model architectures in MRC tasks exceeds the scope of our current study. Nonetheless, we acknowledge that additional factors, such as the specific instruct-fine-tuning of Flan-T5 models on MRC datasets like SQuAD, might also play a role in their enhanced performance over Llama 2 and Alpaca.

\end{document}